\pdfoutput=1
\documentclass[runningheads]{llncs}

\usepackage[utf8]{inputenc} 
\usepackage[T1]{fontenc}    
\usepackage{hyperref}       
\usepackage{url}            
\usepackage{booktabs}       
\usepackage{amsfonts}       
\usepackage{nicefrac}       
\usepackage{microtype}      
\usepackage{xcolor}         
\usepackage{textcomp}
\usepackage{CJKutf8}

\usepackage{booktabs} 
\usepackage{amsmath,amssymb} 
\usepackage{graphicx}
\usepackage{multirow}
\usepackage{soul,color}
\usepackage{bm}
\usepackage{dsfont}
\usepackage{enumitem}
\usepackage{algorithmic}
\usepackage[linesnumbered,ruled,vlined]{algorithm2e}
\usepackage{booktabs}
\usepackage{url}
\usepackage{cite}
\usepackage{tabularx}

\usepackage{wrapfig, blindtext}
\usepackage{etoolbox}
\usepackage{bbm}
\usepackage{float} 
\usepackage{makecell}
\usepackage{setspace}
\usepackage{amsfonts}
\usepackage{subfig}

\newcommand{\synm}{synonymous\ }

\newcommand{\ct}{classroom teaching\ }

\newcommand{\dig}{dialogue}

\begin{document}
\title{Towards Applying Powerful Large AI Models in Classroom Teaching: Opportunities, Challenges and Prospects}
\titlerunning{Opportunities, Challenges and Prospects of LLMs in Classroom Teaching}

\author{Kehui Tan \and Tianqi Pang  \and 
Chenyou Fan \and Song Yu\thanks{Corresponding.}}

\authorrunning{K. Tan et al.}

\institute{South China Normal University, Guangdong, China \\
\email{\{20214001062,2022024954\}@m.scnu.edu.cn, fanchenyou@scnu.edu.cn, sungyuepku@foxmail.com}}

\maketitle              
\begin{abstract}

This perspective paper proposes a series of interactive scenarios that utilize Artificial Intelligence (AI) to enhance classroom teaching, such as dialogue auto-completion, knowledge and style transfer, and assessment of AI-generated content. By leveraging recent developments in Large Language Models (LLMs), we explore the potential of AI to augment and enrich teacher-student dialogues and improve the quality of teaching. Our goal is to produce innovative and meaningful conversations between teachers and students, create standards for evaluation, and improve the efficacy of AI-for-Education initiatives. In Section~\ref{sec:model}, we discuss the challenges of utilizing existing LLMs to effectively complete the educated tasks and present a unified framework for addressing diverse education dataset, processing lengthy conversations, and condensing information to better accomplish more downstream tasks. In Section~\ref{sec:tasks}, we summarize the pivoting tasks including Teacher-Student Dialogue Auto-Completion, Expert Teaching Knowledge and Style Transfer, and Assessment of AI-Generated Content (AIGC), providing a clear path for future research. In Section~\ref{sec:fine-tune}, we also explore the use of external and adjustable LLMs to improve the generated content through human-in-the-loop supervision and reinforcement learning. Ultimately, this paper seeks to highlight the potential for AI to aid the field of education and promote its further exploration.

\end{abstract}
%
%
%
%

%
%
%

\section{Introduction}

\begin{figure}[ht]
\centering
\begin{center}
\centering
\includegraphics[width=0.9\textwidth]{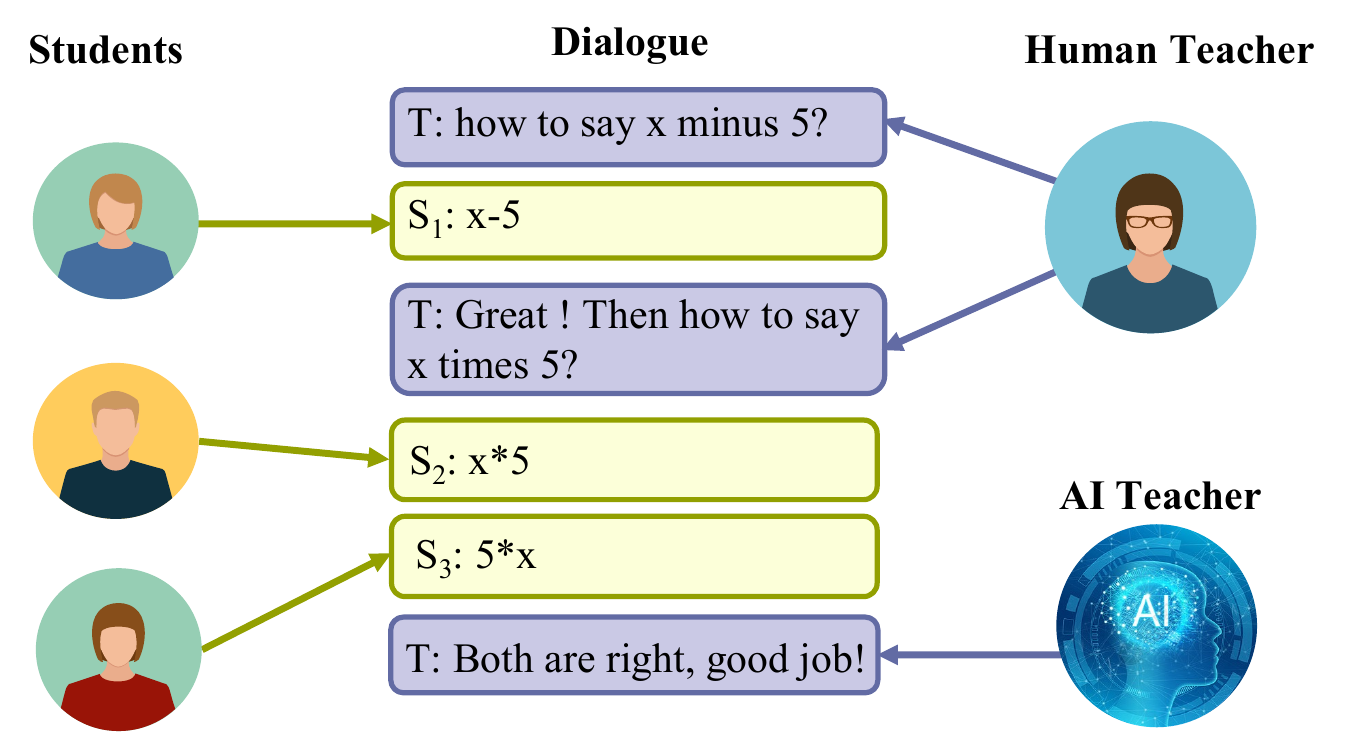}
\end{center}
\caption{An overview of classroom teaching dialogue. In our AI-aimed teaching scenarios, the teacher's conversations can be generated either by humans or AIs. }
\label{fig:Introduction_image}
\vspace{-8pt}
\end{figure}


Classroom teaching is a crucial social process that involves extended interactions between teachers and students. Within this context, teacher-student dialogues promote the exchange of knowledge, skills, and ideas, and demonstrate cognitive processes such as questioning, critical thinking, answering, reflecting, and receiving feedback. These dialogues serve as a vital medium for effective communication and the development of independent thinking skills among students.

In this perspective paper, we explore the use of Artificial Intelligence for Education and focus on 
utilizing the recent successful Large Language Models (LLMs) for
improving classroom teaching by generating meaningful and creative dialogues between teachers and students and providing benchmarks for evaluation.


The recent advancement of large language models (LLMs), such as the GPT series~\cite{brown2020language,ouyang2022training}, has greatly transformed both academic and practical approaches to natural language processing (NLP) tasks. Examples of these tasks include conversational querying~\cite{ouyang2022training}, sentiment analysis~\cite{wankhade2022survey}, and question answering~\cite{zaib2022conversational}. LLMs also exhibit impressive capabilities in generating contextually rich dialogues and demonstrating logical deduction skills. In fact, LLMs have been proven to provide accurate reasoning steps for even complex arithmetic questions, surpassing the ability of human experts in certain cases~\cite{openai2023gpt4}. The AI-generated content is usually abbreviated as AIGC.

In typical primary and high school classrooms, teachers interact with classes of twenty to forty students through dialogues while following a prepared teaching scheme. 
Table~\ref{tab:dialogue_demo} illustrates two exemplary dialogues as transcriptions from realistic teaching activities in a Chinese primary school.
Example 1 shows that in arithmetic class, primary school students understand  ``Why $0$ times $5$ equals to $0$?''. Example 2 shows in Chinese literature class, the students appreciate the artistic conception of Tang poetry and learn classical characters. 
In this article, \emph{we envision the possibility of applying LLMs towards improving the \ct quality and creating AIGC to replenish the teaching scheme database.}


The process of dialogue between teachers and students is not always flawless. One factor that can disrupt the procedure is the occurrence of unforeseen events. For instance, technical malfunctions, such as a sudden breakdown of the projector, can impede the flow of the dialogue due to the lack of presentation tools. This can be challenging for human teachers to quickly revise the dialogue effectively and is referred to as a case of \textbf{``auto-completion.''} 
\begin{CJK*}{UTF8}{gkai}
\begin{table}[t]
    \centering 
    \vspace{-5pt}
    \caption{Dialogue examples for arithmetic and Chinese literature.}
    \begin{tabular}{ll}
         \toprule
         \emph{\textbf{Example 1 (English translation):}}&\emph{\textbf{Example 1 (Chinese):}}\\
         \midrule
         \ \ \ \ \ \ \ \ \ \ \ \ \ \ \ \ \ \ \ \ \ \  $...$ &\ \ \ \ \ \ \ \ \ \ \ \ \ \ \ \ \ \ \ \ \ \  $...$ \\
         \emph{\small \textbf{T:} What $0$ times $5$ equals to, and why?}&\emph{\textbf{T:} $0×5$等于多少，为什么？}\\
         \emph{\small \textbf{$S_1$:} $0$ times $5$ is equal to $0$. Because $3$ }&\emph{\textbf{$S_1$:} $0×5$等于$0$，因为$3×5$等于三个五}\\
         \emph{\small \ \ \ \ \ times $5$ is $15$, $2$ times $5$ is $10$, and $1$ }&\emph{\ \ \ \ \ 相加等于十五，$2×5$等于两个五相}\\
         \emph{\small \ \ \ \ \ times $5$ is $5$. So, $0$ times $5$ means}&\emph{\ \ \ \ \ 加等于十，$1×5$ 等于一个五等于五，}\\
         \emph{\small \ \ \ \ \ there are no fives, which equals $0$.}&\emph{\ \ \ \ \ 所以$0×5$就是没有五等于$0$。}\\
         \emph{\small \textbf{T:} Well said. Are there any other methods?}&\emph{\textbf{T:} 说得真好，还有其他方法吗？}\\
         \emph{\small \textbf{$S_2$:} $0$ times $5$ is equal to $0$, which means $5$}&\emph{\textbf{$S_2$:} $0×5$等于$0$，表示$5$个$0$相加等于$0$。}\\
         \emph{\small \ \ \ \ \ zeros added together equals $0$.}\\
          \ \ \ \ \ \ \ \ \ \ \ \ \ \ \ \ \ \ \ \ \ \ \textbf{$...$} &\ \ \ \ \ \ \ \ \ \ \ \ \ \ \ \ \ \ \ \ \ \  \textbf{$...$} \\
         \ \ \ \ \ \ \ \ \ \ \ \ \ \textcolor{red} {（ Token: 4,312 ）}&\ \ \ \ \ \ \ \ \ \ \ \ \ \textcolor{red} {（ Token: 10,043 ）}\\
         \toprule
         \emph{\textbf{Example 2 (English translation):}}&\emph{\textbf{Example 2 (Chinese):}}\\
         \midrule
          \ \ \ \ \ \ \ \ \ \ \ \ \ \ \ \ \ \ \ \ \ \  \textbf{$...$} &\ \ \ \ \ \ \ \ \ \ \ \ \ \ \ \ \ \ \ \ \ \  \textbf{$...$} \\
         \emph{\small \textbf{T:} What things are mentioned in this poem?}&\emph{\textbf{T:} 这首词都出现了哪些事物？}\\
         \emph{\small \textbf{$S_1$:} The upper part of the poem mentions wine,}&\emph{\textbf{$S_1$:} 上篇出现的词有酒，天气，旧亭台,}\\
         \emph{\small \ \ \ \ \ weather,an old pavilion, and the setting sun.}&\emph{夕阳。}\\
         \emph{\small \textbf{T:} Okay, what words are mentioned in the lower}&\emph{\textbf{T:} 好的，那下篇了出现了哪些词？}\\
         \emph{\small \ \ \ \ \ part of the poem?}\\
         \emph{\small \textbf{$S_2$:} The word ``falling flowers and returning wild}&\emph{\textbf{$S_2$:} 下篇出现了落花归雁。}\\
         \emph{\small \ \ \ \ \  gees'' is mentioned in the lower part of the}\\
         \emph{\small \ \ \ \ \ poem.}\\
         \emph{\small \textbf{T:} Correct, returning geese.}&\emph{\textbf{T:} 对的归雁。}\\
          \ \ \ \ \ \ \ \ \ \ \ \ \ \ \ \ \ \ \ \ \ \  \textbf{$...$} &\ \ \ \ \ \ \ \ \ \ \ \ \ \ \ \ \ \ \ \ \ \ \textbf{$...$} \\
          \ \ \ \ \ \ \ \ \ \ \ \ \ \textcolor{red} {（ Token: 4,102 ）}&\ \ \ \ \ \ \ \ \ \ \ \ \ \textcolor{red} {（ Token: 10,946 ）}\\
        \bottomrule
    \end{tabular}
    \label{tab:dialogue_demo}
    \vspace{-10pt}
\end{table}
\end{CJK*}

Additionally, there may be times when the class syllabus is temporarily altered, requiring the teacher to design a new lesson within a limited time frame for public demonstrations or other events. Consequently, it can be difficult to ensure the delivery of a quality and effectively designed lesson, known as a case of \textbf{``style transfer.''} Another challenge is faced by novice teachers who may struggle to develop an effective dialogue plan, given their relative lack of experience and limited access to teaching resources such as senior teachers' templates. This is known as a case of \textbf{``knowledge transfer.''} 

Finally, The task of fairly grading teaching dialogues and offering constructive feedback to teachers for self-enhancement remains a challenging issue in pedagogy that presently necessitates meticulous human evaluations. This conundrum is referred to as \textbf{``dialogue assessment.''}


In this perspective paper, we propose that advanced AI models can play a crucial role in addressing key interactive scenarios during emergency situations and improving the overall quality of dialogue. Specifically, based on the annotated cases mentioned above, we identify three crucial and imperative areas of research: Teacher-Student Dialogue Auto-Completion, Expert Teaching Knowledge and Style Transfer, and Assessment of AIGC. By integrating LLMs and AIGCs into classroom teaching, we aim to address the opportunities and challenges that come with this innovative approach. We will elaborate on these scenarios in Section~\ref{sec:tasks}.


Additionally, we propose a comprehensive evaluation framework for AIGC used in classroom teaching. This framework incorporates three evaluation methods that have received little attention in previous studies: human-feedback evaluation (HF-Eval), evaluation by language models (LLM-Eval), and evaluation by external language models (Ext-LLM-Eval). The HF-Eval method involves obtaining scores from senior human teachers, which is a specialized but expensive and non-scalable approach. To address this, we suggest the LLM-Eval method, which uses a pre-designed prompting template for self-evaluation but may lack subject-specific knowledge or language context. Finally, for the Ext-LLM-Eval method, we propose using a specialized external LLM to assess the teaching material database. Based on these grading methods, we suggest two practical ways of fine-tuning the Ext-LLMs to generate improved teaching-related content.

We summarize our contributions in this perspective paper as follows. 
\begin{enumerate}[noitemsep,leftmargin=*]

\item We consider a critical and broad topic in education that how to incorporate the capable large language models into \ct properly and effectively.

\item We envision three most applicable teaching scenarios within the classrooms, including AIGC for teaching scheme, style transfer, and auto-grading.

\item We show self-collected data with literature, elementary arithmetic, and reasoning, which we would like to boost future research.

\item We identify several practical ways of grading the quality with AI, which can be typically useful for classroom teaching improvement.

\item We propose a novel way of using external LLM for evaluating and fine-tuning.

\end{enumerate}

\section{Related Work}


\textbf{Large Language Models (LLMs).}
Since the inception of Transformer architecture~\cite{vaswani2017attention} in 2007,  Transformer-based language models such as ELMo~\cite{Peters2018DeepCW} and BERT~\cite{devlin2018bert} have dominated the NLP realms by producing leading performance. 

Recently, the practice of scaling up model sizes for building large language models (LLMs) has received substantial attentions. The proposed LLMs such as the GPT family~\cite{brown2020language,ouyang2022training}, 
PaLM~\cite{chowdhery2022palm}, Galactica~\cite{GALACTICA} , LaMDA~\cite{thoppilan2022lamda}, LLaMA~\cite{touvron2023llama} have been emerging with enormous parameters from 10B to 1000B parameters. These huge models show great capacities in understanding human languages and performing reasoning tasks such as arithmetic reasoning and question answering.
Inspired by LLMs, Large Visual Models (LVMs), such as Vision Transformer (ViT)~\cite{dosovitskiy2020image} and SAM~\cite{kirillov2023segany}, and multi-modality models, such as CLIP\cite{radford2021learning}, have also emerged. These studies propose various ways of fusing different modalities such as texts~\cite{devlin2018bert}, image patches~\cite{chen2020uniter,radford2021learning}, segmentation masks~\cite{oquab2023dinov2}, and sounds~\cite{radford2022robust}.

Many recent efforts have been made for improving the LLMs answering quality. The \emph{Chain-of-Thought} (CoT)~\cite{wei2022chain} technique proposes to add existing intermediate reasoning steps as hints at the head of the original query, which has substantial improvement on reasoning tasks~\cite{zhang2022automatic}. The \emph{Self-consistency}~\cite{wang2023selfconsistency} technique proposes to develop multiple reasoning paths followed by making a majority vote to explore the most likely answer. The Federated-LLM~\cite{liu2023fed} proposes to improve distributed query answering by mining \synm questions.  These progressives are extensively discussed in recent  surveys~\cite{zhao2023survey,yang2023harnessing,cao2023comprehensive}.

\textbf{AI-Generated Content (AIGC)}.  Generating virtual contents with combinations of LLMs and LVMs has been emerging and gained much research and commercial attentions. Different from human generated content, seemingly genuine AIGC is generated automatically with tremendous speed by Generative AI (GAI) models, such as GPT-4~\cite{openai2023gpt4}, DALL-E2~\cite{ramesh2022hierarchical}, Codex~\cite{Chen2021EvaluatingLL}. The generated creative contents range from texts~\cite{openai2023gpt4}, captions~\cite{li2023blip} to artistic images~\cite{ramesh2022hierarchical}, videos~\cite{midjourney2023}, financial insights~\cite{wu2023bloomberggpt} and programming code~\cite{Chen2021EvaluatingLL}, etc.


\textbf{Teacher-student \dig \ } is the main media of interaction in classroom teaching and has been extensively studied in the pedagogical field. 
Numerous studies have focused on teaching dialogues across various disciplines, including mathematics and linguistics~\cite{loewen2018interaction, masatoshi2016understanding, mercer2008seeds}. Some studies have provided evidence that science and math courses benefit more from effective dialogues~\cite{asterhan2015socializing, song2019exploring}. Additionally, efforts have been made to improve student engagement in classroom learning through dialogue, as evidenced by several studies~\cite{howe2013classroom, howe2019teacher, song2019exploring, song2021automatic}.

\textbf{Dialogue assessment.}
Assessing classroom dialogues has been a significant area of focus in pedagogical research, particularly in the fields of mathematics and science~\cite{howe2017commentary,alexander2001border,asterhan2015socializing}. Various approaches have been employed, such as collecting student feedback for detailed evaluation~\cite{sinclair2013towards,schwab2011dialogue}, and using machine learning methods for prediction~\cite{song2021automatic,song2021comparative}. Some studies have evaluated the students' thinking process from diverse points of view~\cite{mercer2014study}, while others have focused on dialogic practices in achieving specific goals~\cite{guzman2021transformation} or the performance of student group interactions~\cite{howe2017commentary} and actual performance scoring~\cite{howe2015principled}.
\emph{However, the generation of dialogues using automated AI tools and the assessment of their quality with human-like standards remains an under-explored area.}



The following sections of this paper are organized as follows: Section~\ref{sec:model} provides an overview of how to condense lengthy teaching dialogues into dense representations and integrate LLMs to carry out downstream tasks. In Section~\ref{sec:tasks}, we offer detailed explanations of three prevalent scenarios in classroom teaching that AIGC can enhance. Section~\ref{sec:fine-tune} outlines a novel approach to refining the capability and interpretability of an external LLM.



\section{Model Architecture}
\label{sec:model}

\begin{figure}[ht]
\centering
\begin{center}
\includegraphics[width=0.8\textwidth]{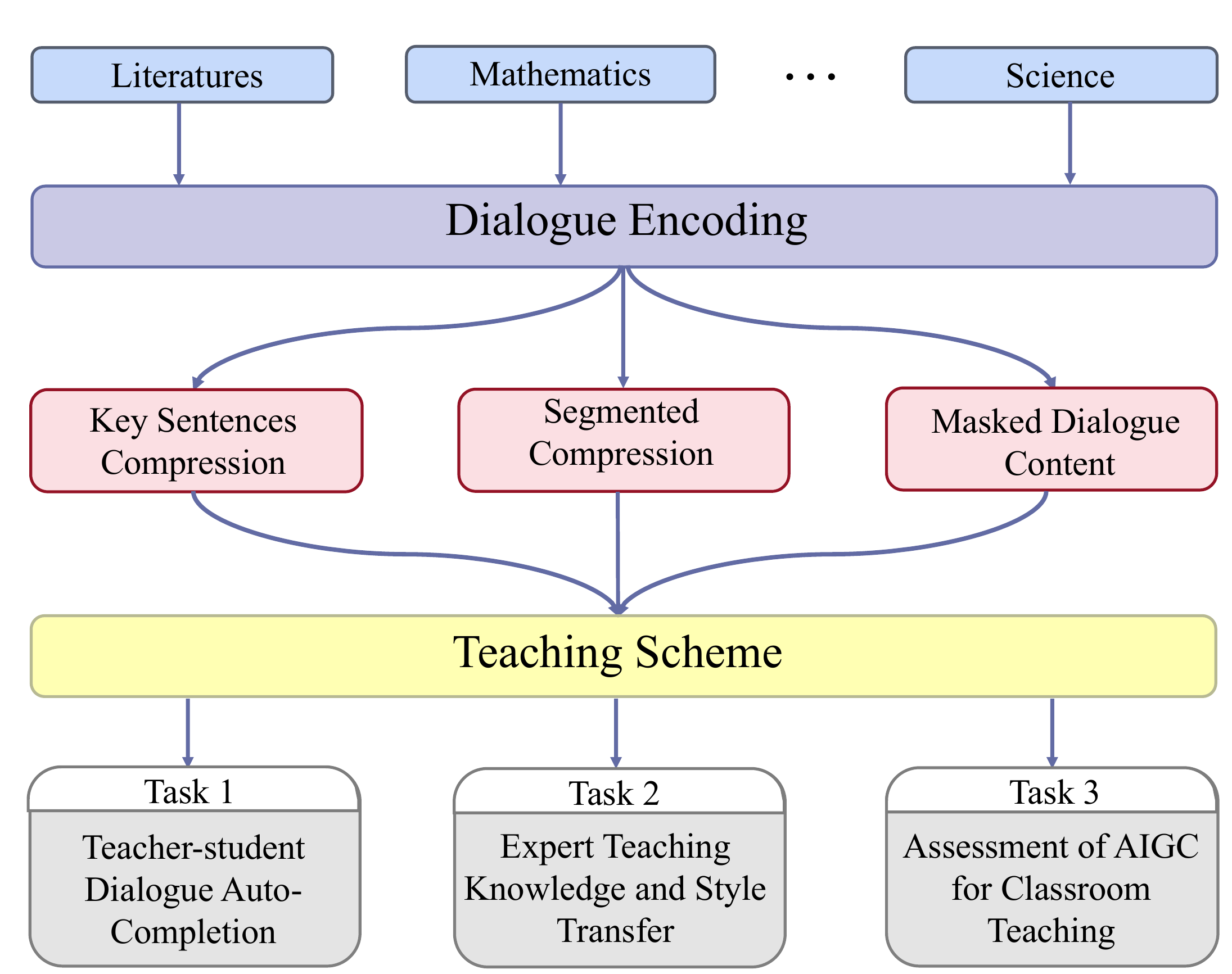}
\end{center}
\caption{Illustration of a unified framework for achieving the AIGC in classroom teaching. It takes various practical teaching topics and encodes the dialogues into summarized content. This content is then fed to LLMs to generate output for a variety of downstream tasks, such as auto-completion and knowledge transfer. 
}
\label{fig:approach_pipeline}
\vspace{-10pt}
\end{figure}

In this section, we discuss the challenges of utilizing existing LLMs to effectively generate teaching dialogues and present a unified framework as in Fig.~\ref{fig:approach_pipeline} for addressing diverse teaching subjects, processing lengthy conversations, and condensing key information for subsequent analysis.

\subsection{Segmented Text Encoding}
Standard conversational chatbots like GPT-3.5~\cite{ouyang2022training} follow a query-answer mechanism. However, this approach is insufficient for properly understanding a lengthy classroom teaching dialogue as it almost always surpasses the token limits, compromising the understanding of key content and leading to poor downstream tasks like dialogue completion. We make a detailed explanation below.

\begin{enumerate}[noitemsep,leftmargin=*]
\item As shown in Table.~\ref{tab:dialogue_demo}, a typical 40-minute classroom teaching session usually composes of about 120$\sim$140 rounds of teacher-student spoken sentences. Considering that each sentence usually consists of 10$\sim$20 words, the entire dialogue encompasses nearly 6,000 words, accounting for over 12,000 tokens. However, the maximum limit of input token for a typical LLM such as GPT-3.5 is only 4,096 tokens, indicating its inability to accept such lengthy textual input directly. Consequently, the system fails to produce desirable outputs as per our requirements.

\item Although the conversations may be concise enough to fit in the LLM, the frequent role changes between student and teacher can have a significantly adverse effect on the LLM's ability to accurately comprehend the dialogue. This can lead to poor performance on downstream tasks, such as question-answering or dialogue completion. For instance, we have experimented and observed that attempting to complete an unfinished dialogue directly often results in yet another iteration of the previous round of the teacher's question being repeated. Obviously, it is not the effect we are expecting.
\end{enumerate}





\textbf{To address these challenges}, we propose a unified framework that can effectively summarize the input dialogue into a condensed teaching scheme, as shown in Fig.~\ref{fig:approach_pipeline}. The core module of this framework is the Dialogue Encoder (D-Encoder), which is specifically designed to compress the conversations into dense representation that we call \emph{``Teaching Scheme’'}. The Teaching Scheme is supposed to fully maintain useful information to facilitate subsequent tasks.

Hereby, the first key research question is how to fulfill the D-Encoder in an effective way. We propose the following feasible methods to address this question.
\begin{enumerate}[noitemsep,leftmargin=*]
\item \textbf{Key Sentences Compression}. For a well-constructed teaching database, each dialogue should be labeled with fine-grained labels indicating the progression of the dialogue. This would enable us to automatically extract the key sentences and merge them into a prompt, such as ``Summarize the instructional approach of this class based on the aforementioned key phrases.'' Language models can then generate a well-informed teaching scheme based on the given context in the prompt. However, tagging the dialogues requires human expertise and is thus costly and time-consuming.
\item \textbf{Segmented Compression}. The second method we consider is to divide the dialogue by logical stages of the class. Several existed studies showed the feasibility of providing sentence-level logic labels based on contexts~\cite{song2021automatic}. Thus, we can utilize LLMs to summarize each segmented content separately and merge all parts to obtain the Teaching Scheme.
\item \textbf{Masked Dialogue Content}. We propose a teacher-centered approach to focus mainly on teacher's guidance in a dialogue.  A recent study~\cite{wei2022chain} shows that forcing the LLMs to output sequence of dots '...' cannot stop the innate thinking process of the LLMs. We hypothesize that replacing student inputs with '...' would not significantly impact the overall dialogue content. Preliminary results suggest that language models are capable of making accurate predictions of student responses even when their true responses are masked. Therefore, we propose the use of masked dialogue to filter out student utterances such as ``hmmmm'' or ``uh-huh''. This practice can substantially reduce the number of tokens used as inputs to language models, while still preserving maximum dialogical information. We even speculate that the masked dialogue content could potentially lead to improved summarization, as it eliminates confusing and fragmentary utterances.
\end{enumerate}

By utilizing our proposed D-Encoder to condense an extensive dialogue, the LLMs have a higher likelihood of comprehending the primary information and providing more detailed and concise Teaching Schemes for future tasks. The Teaching Scheme can then be utilized in various teaching scenarios, and we will further elaborate on these scenarios in the subsequent section.


\section{Scenarios and Approaches}
\label{sec:tasks}

In this section we present the realistic scenarios that can be seamlessly combined with AIGC to assist teachers in their classroom activities. Our proposed approaches aim to enhance the quality of instructional dialogues and facilitate constructive feedback for teachers. We elaborate these scenarios as follows.

\subsection{Teacher-student Dialogue Auto-Completion}



The first practical user scenario we consider is generating missing parts of teacher-student dialogue, termed as \emph{Teacher-Student Dialogue Auto-Completion}. This consideration is motivated by the potential disruptions that can occur during classroom teaching. Technical malfunctions, such as a sudden breakdown of a projector or the teacher having to tend to an ill student, can interrupt the flow of dialogue. In such cases, an AI system must rapidly provide an alternate plan for dialogue to continue classroom teaching. This presents a significant challenge for the AI system.

Depending on the task setting, we identify two auto-completion tasks: 1) Generating the complete teaching dialogue through a given teaching scheme or description, and 2) Supplementing the incomplete excerpt of the dialogue based on its context within the dialogue.


For the first case of \textbf{generating the entire teaching dialogue} through teaching scheme, we can add the suffix ``Please write a classroom dialogue between teacher and students based on the above teaching scheme (no less than 15 rounds)'' as a prompt, to expand the abstract to a concrete dialogue with detailed teacher-student conversations.


For the second case of \textbf{supplementing an incomplete dialogue}, we leverage the powerful masked language generation capability of LLMs by requesting the missing dialogical portion. Our research has shown that a straightforward prompting technique is adequate for this purpose. We designate the conversations or Teacher Scheme prior to the missing portion as Part-1, and those following it as Part-3. Our prompt is as follows: ``Please read the Part-1 and Part-3 dialogues, and generate the missing Part-2 to complete the dialogue. [Part-1]. [Part-3].'' 


Besides the aforementioned prompt-based auto-completion methods, we provide two further options to pinpoint specific tasks.

\textbf{Utilizing External Specialized APIs}. To accomplish specialized tasks, it may be necessary to utilize external APIs that can ensure the accuracy of the required result. For example, the Wolfram API~\cite{wolfram2023} can be utilized for calculating trigonometric functions that are part of the curriculum material provided by the instructor. Similarly, BloombergGPT~\cite{wu2023bloomberggpt} can be used for better comprehension of financial courses.

\textbf{Dynamic Role Setting}.
We can dynamically allocate the roles according to the subject to enhance the performance of dialogue completion. We can prompt the LLMs with a specific role, such as ``Imagine you are a senior \textbf{[ROLE]} and complete the classroom dialogue based on the given context.'' This role could be filled by a variety of professionals, such as English literature, mathematics, or physics teachers accordingly. This explicit setting of role player provides specific context for the generated content.  


\subsection{Expert Teaching Knowledge and Style Transfer}
\label{sec:teaching_style}

The second practical user scenario we consider is as follows. Given a few-shot set of classroom teaching dialogues from senior instructors, \emph{can we create more professional dialogues for different topics or subjects, through prompting AI models to perform knowledge and style transfer?}


It is universally acknowledged that developing a teacher's pedagogical abilities necessitates constant emulation of successful teaching methodologies in improving their own teaching proficiency. However, there is a scarcity of expert classroom recordings accessible in the teaching dialogue repository and the inconsistency between the content taught by teachers and that covered in the expert recordings. 
Manually transferring a teaching dialogue to another topic can be a difficult task for human teachers. Firstly, they may not have a full grasp of the original dialogue, and this could be further complicated when attempting to transfer it to a new topic. Secondly, even if they have a thorough understanding of the original dialogue, they would need to possess an equal level of expertise in the new subject in order to successfully transfer it, which is a challenge even for senior instructors.


Therefore, our task is to leverage the superior understanding capacity of large AI models to comprehend the existing expert teaching dialogues, then  adequately extending to a new topic with the equal-level teaching wisdom, similar methodology and closed fashion. 
Using AI to imitate the teaching styles of experts can produce a wealth of high-quality classroom dialogues useful for learning and referencing. Hence, creating additional teaching plans and classroom dialogue material in line with the given expert content would be extremely worthwhile and advantageous.

Depending on the task setting, we  discuss two transfer tasks and proposed methods as follows.

1) \textbf{Zero-shot Transfer}: Generating a teaching dialogue from scratch given a topic description only, e.g., ``Instruct Newton's Laws of Motion step-by-step.''

For this task, we propose the following transfer procedures.
Initially, we conduct semantic matching with public or specialized proprietary teaching dialogue databases. The objective is to directly match credible teaching dialogues that already exist, if a perfect match can be found. This is especially useful for extensively researched topics such as literature and arithmetic, for instance, comprehending Li Bai's ``Quiet Night Thoughts'' or Newton's Laws of Motion in classical physics. Nevertheless, it’s unlikely that the perfect match would be obtainable for all topics and subjects. In such cases, we can generate a set of best-matching templates from the database as a single Chain-of-thought (CoT). The entire CoT is appended as a prefix to the user's query prompt in order to obtain the final answer by the LLMs. We can provide an error disclaimer~\cite{liu2023fed} at the end of this prompt, stating that ``the provided template styles may not be fully applicable, please think more carefully.'' to gain better results.


2) \textbf{Template Transfer}: Generating a teaching dialogue for a new subject from a concrete dialogue template. An example is to present a dialogue about teaching \emph{merge-sort} and then use AI to adapt the underlying \emph{divide-and-conquer} idea to teach \emph{quick-sort}, and finally, prompt a dialogue.

To address the task of template transfer, we can employ a similar semantic matching technique with the input template to establish more precise matching conditions. Subsequently, we can produce a CoT template for generating transferred dialogues. However, we differentiate template transfer from zero-shot transfer by adding a claim that the provided template is precisely the style we intend to transfer. Additionally, we include a disclaimer to emphasize that the given template is more critical than database-retrieved ones. 

\subsection{Assessment of AIGC for Classroom Teaching}
\label{sec:grading}
The third practical user scenario we consider is to assess the AIGC for classroom teaching fairly and consistently. The crucial challenge 
for using AIGC directly in classroom setting is  how to evaluate the excellence of the content generated by the AIGC. To overcome this challenge, we propose a set of strategies to promote equitable and competent grading and comparable to human expert evaluations.




1) \textbf{Grading Based on Human-Feedback.}
We may seek the input of human pedagogical experts and senior teachers to evaluate the quality of our generated dialogues and content.
For example, a usual practice is to ask the human evaluators to assess the quality of the generated answer on a scale of 1 to 10, with 1 indicating the lowest quality and 10 indicating the highest quality, and to provide a written explanation justifying their scores.
To avoid personal biases, we ensure that each teaching case is scored by multiple experts and calculate the final score for each case based on a reasonable set of scores or majority voting. 

The outcomes from this process are then averaged to obtain a final grade for each piece of content. Nevertheless, it is evident that this Human-Feedback based rating approach has a crucial drawback: it can be very costly since it requires human involvement. As a result, it cannot be easily scaled up in the same way as AIGC, and it is limited in its applicability.

2) \textbf{Grading Based on LLMs.}
LLMs are versatile and capable AI models that, theoretically, can grade their own answers effectively. To achieve this self-grading ability, we can follow a methodology called Self-Ask~\cite{press2022measuring}.


Specifically, in order to evaluate the quality of generated answers for down-stream tasks, we utilize a prompting prefix to indicate response was self-generated based on the preceding conversation. Then we can obtain LLM's grading and comments accordingly. By automating this process using the prompting prefix, we can efficiently obtain a comprehensive set of scores for all generated answers.

Nevertheless, self-grading has systematic drawbacks. First, the general-purpose LLMs lack the background knowledge necessary to accurately assess educational objectives. As a result, grading answers is often based on general guesses such as the fluency of the dialogue, rather than on dialogical reasons. Additionally, it is widely known that most LLMs are better suited to understanding English contexts, as the majority of training corpora are in English. Therefore, using LLMs to grade Chinese poems and dramas in the classroom may be unreliable.




3) \textbf{Grading and Commenting Based on Ext-LLMs.} 
Considering the drawbacks associated with human evaluators or non-specialized Language Model Models (LLMs) when it comes to grading, we provide an innovative and potential approach. This approach involves training a medium-sized External LLM (Ext-LLM) for the purpose of evaluating and providing feedback.

The primary consideration when selecting this Ext-LLM is that it should be open-source and come with pre-trained models. This allows us to have full control over the storage, fine-tuning, and deployment of the model. In contrast, commercial LLMs such as GPT only provide private and paid APIs.
The aim is to ensure the representative power of the Ext-LLM while still allowing for the flexibility of fine-tuning and deploying with hardware requirements that are reasonable and affordable for academic institutes.


Now we explain how such an Ext-LLM can be tuned for grading purpose.

Inspired by the \textbf{Reward Modeling (RM}) technique used for training GPT-3.5~\cite{ouyang2022training},
we formulate the grading task as a supervised regression with ground-truth human evaluated grades, based on given dialogue as the contexts.
By comparing the AI-predicted and human evaluated ratings, the Ext-LLM can be adjusted as a reliable scoring model with human expertise from reward modeling as well as language understanding capability from pre-trained parameters with natural language understanding tasks.

We can naturally extend the grading functionality to commenting the generated dialogues. As generating comments is indeed another AI-generation task, we assume that the Ext-LLM has already gained the commenting capability either from pre-trained model or fine-tuning with realistic human evaluations.

We can apply the Chain-of-Thought (CoT) method to guide the comment generation to ensure they are close to human expert evaluations.
To achieve this, we can attach a suffix to the generated content ``Therefore, assuming you are an elite teacher, please provide a professional evaluation given the dialogue.'' to provide the final comments. This procedure is also termed zero-shot CoT that was initially used to improve reasoning performance~\cite{brown2020language}. We can also employ the more effective few-shot CoT~\cite{brown2020language} by giving realistic human evaluations as prefix of the prompt additionally. This has also been discussed in Section~\ref{sec:teaching_style}.





\section{AI Models Fine-tuning with Teaching Databases}
\label{sec:fine-tune}


In this section, we outline our ideas for refining the capability and interpretability of large AI models in order to unlock their domain-specific capabilities.

\label{sec:rw}
\textbf{Supervised ranking loss.} Inspired by the supervised Reward Modeling methodology as mentioned in Section~\ref{sec:grading}, we can design the following contrastive ranking loss to align the AI-generated grading with human evaluations. Specifically, we can randomly select $K$ AI-generated answers in each round. We then ask both the human annotators and the Ext-LLM to grade the answers based on the same contexts for generation. Thus, we obtain $K$ pairs of human and AI grading as a dataset $\mathcal{D}$.

We design a ranking loss function to align the Ext-LLM grading capacity with humans by minimizing the discrepancy of their grading results, as follows:
\begin{equation}
\label{eq:rank_loss}
\operatorname{loss}(\theta)=-\frac{1}{K(K-1)/2} E_{\left(x, y_1, y_2\right) \sim D}\left[\log \left(\sigma\left(r_\theta\left(x, y_1\right)-r_\theta\left(x, y_2\right)\right)\right)\right] \ .
\end{equation}

In Eq.~\ref{eq:rank_loss}, the $r_\theta(x, y_{1})$ and $r_\theta(x, y_{2})$ are the Ext-LLM's grading of AI-generated answer $y_1$ and $y_2$ with context $x$, respectively. Notedly, $y_1$ is the preferred answer of the randomly sampled pair $(y_1,y_2)$ by human grading. By minimizing ranking loss~\ref{eq:rank_loss} with $K(K-1)/2$ distinct pairs enumerated from $K$ answers of $\mathcal{D}$, we can fine-tune the Ext-LLM model parameter $\theta$ towards aligning its grading ability with humans. It is convenient to extend such grading alignment procedures to comments by using the reward model for comments.

\textbf{Reinforcement Learning from Human Feedback methodology} (RLHF). Many schools have accumulated self-collected teaching databases over the years. To fully explore those specialized educational datasets to improve AIGC, we can further involve the human-in-the-loop processes for enhancing both assessment and generation. To this end, we can construct more sophisticated and effective approaches by employing extensive human interventions to further refine the Ext-LLM model, based on the RLHF technique~\cite{christiano2017deep,ouyang2022training}.

We take the dialogue auto-completion task as an example.
Given the context and the generated dialogue $y_l$ by LLMs, we train the Ext-LLMs to refine the dialogue as a normal prompt to obtain the refined dialogue $y_e$. 
Let the reward of Ext-LLM's output $y_e$ be $r_\theta(x, y_e)$, and let human annotated grade be $r_h(x, y_e)$. We can utilize the standard RL method such as PPO~\cite{schulman2017proximal} to follow the trace of maximizing the reward $r_\theta(x, y_e)$ while taking human reward as a baseline to make training process robust. During training, we can also minimize the ranking loss~\ref{eq:rank_loss} together as a multi-task learning objective.


\section{Conclusion and Future Challenges} 


In this perspective paper, we investigate the potential paths of applying powerful large AI models in
classroom teaching scenarios, making it one of the earliest efforts to tailor AI-generated content for pedagogical purposes. We start by reviewing recent developments in AIGC, covering existing large AI models and related advancements over the past two years. In Section~\ref{sec:model}, we focus on integrating AIGC into classroom teaching by providing a unified general framework of abstracting prolonged teaching dialogues into dense representations and integrating LLMs to carry out downstream tasks.



In Section~\ref{sec:tasks}, we presented several exemplary classroom teaching dialogues in both English and Chinese for several different courses, including literature, elementary arithmetic, and reasoning. We showed that AIGC could potentially be perfectly fit into practical teaching in several distinct but connected ways, greatly enhancing and enriching existing teaching materials with knowledge and style transfer. For the auto-completion task, we presented two feasible ways of generating dialogues from user prompts, helping teachers to deal with unforeseen events or necessities of revising the teaching plan within a short period of time. For the teaching dialogue transfer task, we showed the potential of AIGC for imitating expert teaching dialogues and providing high-quality dialogues for a wider range of topics and subjects.



In Section~\ref{sec:fine-tune}, we propose fully fine-tuning a controllable and competent External LLM (Ext-LLM) for grading and commenting purposes to enhance the interpretability and explainability of black-box large AI models. Therefore, we propose employing medium-sized Ext-LLMs to maintain the capacity of content generation as well as the flexibility and interpretability by tuning on our own teaching databases or direct from human evaluations and dialogue corrections.

Finally, we identify several open problems related to the utilization of AIGC in classroom teaching, which future works can pay special attention to.
\begin{enumerate}[noitemsep,leftmargin=*]
\item How to extend the power of AI to generate multi-modal and multimedia content, such as teaching videos and audios, across multiple disciplines including arts, music, and physical education.
\item How to ensure the privacy of AIGC is a crucial issue that can hinder its widespread use. Studies can focus on preventing the generation of fake faces through malicious DeepFake techniques~\cite{rossler2019faceforensics,tolosana2020deepfakes}. We can also explore the differential privacy technique~\cite{dwork2014the,abadi2016deep,fan2022private} to protect generated personal data and identities.
\end{enumerate}

Ultimately, we hope that this perspective paper can have a constructive impact on helping more AI and pedagogical researchers to understand the frontiers of emerging concepts in LLMs and AIGC and gain insights into the potential of applying these techniques to causes of education. Moving forward, we provide a comprehensive analysis of the limitations and potentials of AI-generation techniques and propose several accessible directions for further research. We wish for the development of AIGC for educational purposes to be safe and meaningful, so that it can be used to its fullest potential without growing uncontrollably.

{
\bibliographystyle{splncs04}
\bibliography{egbib,plm,edu}
}

\end{document}